\documentclass[10pt,twocolumn,letterpaper]{article}

\usepackage{cvpr}              % To produce the CAMERA-READY version

\usepackage[utf8]{inputenc} % allow utf-8 input
\usepackage[T1]{fontenc}    % use 8-bit T1 fonts
\usepackage{hyperref}       % hyperlinks
\usepackage{url}            % simple URL typesetting
\usepackage{booktabs}       % professional-quality tables
\usepackage{amsfonts}       % blackboard math symbols
\usepackage{nicefrac}       % compact symbols for 1/2, etc.
\usepackage{microtype}      % microtypography
\usepackage{colortbl}
\usepackage{multirow}
\usepackage{amsmath} 
\usepackage{graphicx}
\usepackage{subcaption}
\usepackage{pifont} % Access to PostScript standard Symbol and Dingbats fonts
\usepackage{makecell}

\usepackage{colortbl}
\usepackage{booktabs,multirow}

\definecolor{Best}{HTML}{F8C6C6}    % 연한 빨강
\definecolor{Second}{HTML}{FFD8B1}  % 연한 주황
\definecolor{Third}{HTML}{FFF7CC}   % 연한 노랑

\newcommand{\cellbest}[1]{\cellcolor{Best}\textbf{#1}}
\newcommand{\cellsecond}[1]{\cellcolor{Second}\textbf{#1}}

\definecolor{cvprblue}{rgb}{0.21,0.49,0.74}
\def\paperID{18319} % *** Enter the Paper ID here
\def\confName{CVPR}
\def\confYear{2026}

\title{MoRGS: Efficient Per-Gaussian Motion Reasoning \\ for Streamable Dynamic 3D Scenes}

\author{
Wonjoon Lee$^{1}$ \quad 
Sungmin Woo$^{1}$ \quad
Donghyeong Kim$^{1}$ \quad 
Jungho Lee$^{1}$ \quad 
Sangheon Park$^{1,2}$ \quad 
Sangyoun Lee$^{1}$ \\
$^{1}$Yonsei University\\
$^{2}$Electronics and Telecommunications Research Institute (ETRI)\\
{\tt\small \{lwjinkorea, smw3250, 2donghyung87, 2015142131, shpark12, syleee\}@yonsei.ac.kr}\\
{\tt\small \{shpark12\}@etri.re.kr}
}

\begin{document}
\twocolumn[{
\maketitle
\vspace{-20pt}
\begin{center}
    \includegraphics[width=1.0\textwidth]{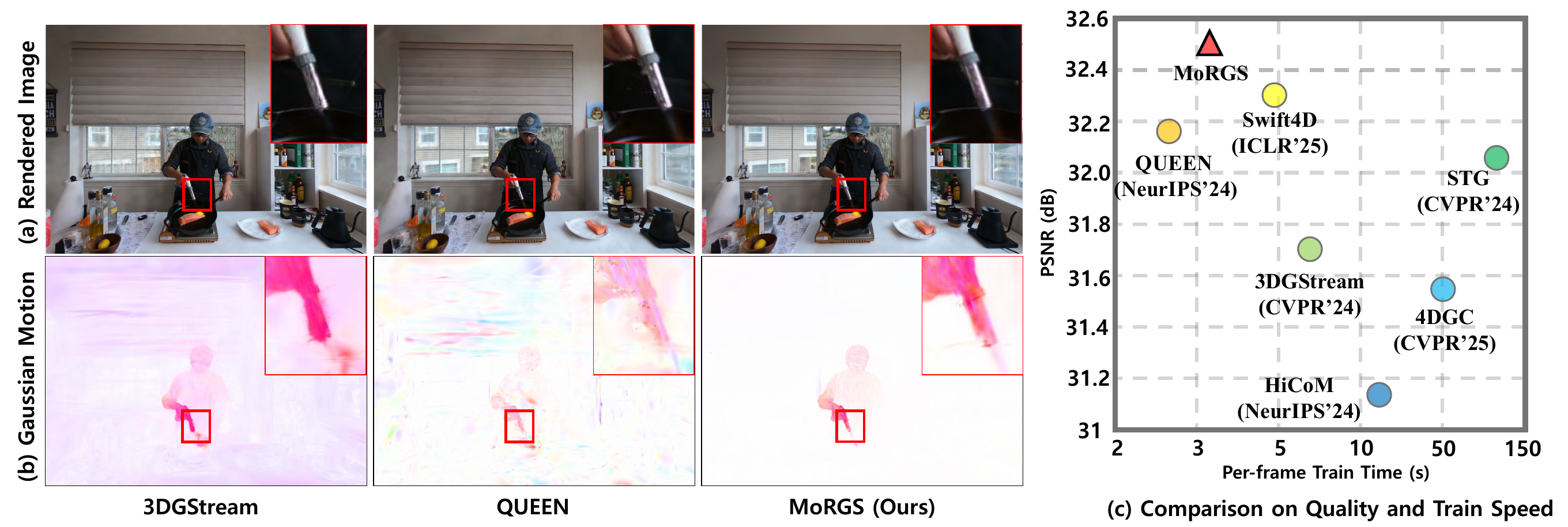}
    \vspace{-10pt}
    \captionof{figure}{
        The proposed MoRGS framework for streamable dynamic scene reconstruction achieves superior rendering quality by explicitly modeling per-Gaussian motion. The left figures ((a),(b)) show the high-quality rendering and the corresponding Gaussian motion updates compared to \cite{3dgstream,queen}. The right figure (c) is the performance comparison with previous state-of-the-art methods~\cite{3dgstream,queen,swift4d,spacetimegaussian,4dgc,hicom}.
    }
\label{fig:teaser}
\end{center}
%\vspace{-5pt}
}] 
\maketitle
\begin{abstract}
Online reconstruction of dynamic scenes aims to learn from streaming multi-view inputs under low-latency constraints. The fast training and real-time rendering capabilities of 3D Gaussian Splatting have made on-the-fly reconstruction practically feasible, enabling online 4D reconstruction. However, existing online approaches, despite their efficiency and visual quality, fail to learn per-Gaussian motion that reflects true scene dynamics. Without explicit motion cues, appearance and motion are optimized solely under photometric loss, causing per-Gaussian motion to chase pixel residuals rather than true 3D motion. To address this, we propose MoRGS, an efficient online per-Gaussian motion reasoning framework that explicitly models per-Gaussian motion to improve 4D reconstruction quality. Specifically, we leverage optical flow on a sparse set of key views as lightweight motion cues that regularize per-Gaussian motion beyond photometric supervision. To compensate for the sparsity of flow supervision, we learn a per-Gaussian motion offset field that reconciles discrepancies between projected 3D motion and observed flow across views and time. In addition, we introduce a per-Gaussian motion confidence that separates dynamic from static Gaussians and weights Gaussian attribute residual updates, thereby suppressing redundant motion in static regions for better temporal consistency and accelerating the modeling of large motions. Extensive experiments demonstrate that MoRGS achieves state-of-the-art reconstruction quality and motion fidelity among online methods, while maintaining streamable performance.
\end{abstract}

\section{Introduction}
\label{sec:intro}
Reconstructing dynamic scenes from multi-view videos enables free-viewpoint navigation across space and time, which is a core capability for AR/VR/XR, telepresence, and immersive media applications. Early neural rendering methods based on Neural Radiance Fields (NeRF)~\cite{nerf} significantly improved 3D reconstruction fidelity and were later extended to dynamic scenes~\cite{nonrigidnerf,dnerf,nerf-ds,streamrf,nerfplayer,rerf,tetrirf}. However, the heavy optimization and slow rendering of NeRF hinder real-time or interactive use. Recently, 3D Gaussian Splatting (3DGS)~\cite{3dgs} has emerged as an alternative that offers faster training and real-time rendering while preserving high visual quality, and it has become a widely used basis for dynamic scene reconstruction~\cite{4dgs,4d-gs,deformabl3dgs,spacetimegaussian,splinegs,rotor4dgs,swift4d,motiongs,fedgs,sarogs}.

However, most dynamic scene reconstruction methods remain offline and require access to the full sequence. While offline models can capture complex spatiotemporal geometry and appearance to achieve high-quality reconstructions, their heavy computation and non-causal training make them incompatible with live streaming applications, where frames arrive sequentially and future observations are unavailable.

Recent works address this gap with online reconstruction methods~\cite{3dgstream,igs,4dgc,dynamic3dgs,hicom,queen} that incrementally estimate inter-frame deformations and update the representation per frame. Online methods enable on-the-fly training and rendering while keeping per-frame memory low for efficient transmission. Despite these strengths, current online approaches fail to learn per-Gaussian motion that reflects true scene dynamics, as shown in Fig.~\ref{fig:teaser}(b). To satisfy strict latency and compute constraints, they avoid using explicit motion cues such as optical flow and instead rely solely on photometric supervision to update both appearance and motion. This creates a mismatch between supervision and objective, where per-Gaussian motion is optimized to reduce pixel residuals rather than to recover actual 3D dynamics. Under this pixel-driven objective, the model explains local appearance changes by slightly displacing nearby Gaussians that should remain static instead of correctly moving the truly dynamic ones. Consequently, Gaussians responsible for large inter-frame motion are underestimated, while static Gaussians acquire redundant motion, leading to degraded temporal consistent reconstruction.

In this paper, we propose MoRGS, an efficient per-Gaussian motion reasoning framework for online dynamic scene reconstruction. Unlike prior online approaches that learn motion implicitly as an appearance-matching proxy, we explicitly model per-Gaussian motion and concentrate updates on truly dynamic Gaussians by efficiently leveraging flow-based motion prior with minimal overhead for online reconstruction.

Specifically, we introduce three components that supervise, refine, and weight per-Gaussian motion. First, we leverage optical flow as an efficient supervisory signal for per-Gaussian motion. Rather than computing dense flow for all views, which would undermine the efficiency of online reconstruction, we compute optical flow only on a sparse set of key views and use it as a lightweight motion cue. For Gaussians contributing to these selected views, we project the per-Gaussian 3D motion onto the image plane and align it with the observed flow, guiding inter-frame updates so that Gaussian motion follows scene geometry instead of merely minimizing photometric error. Second, although flow supervision offers strong directional cues, its sparse and view-limited nature can lead to mismatches between flow and geometry. We therefore learn a per-Gaussian motion offset field that refines the flow-guided motion and reconciles discrepancies between projected 3D motion and observed optical flow, preserving geometric consistency across views and time. Third, to improve temporal consistency and avoid redundant updates, we estimate a per-Gaussian motion confidence that weights Gaussian attribute residual updates. This motion confidence prioritizes updates for Gaussians with reliable motion evidence while attenuating gradients for near-static ones. This selective focus accelerates the modeling of large motions early in training and reduces unnecessary updates in static regions, yielding more temporally stable online reconstructions. Taken together, these components form a unified, motion-aware framework that aligns per-Gaussian updates with true scene dynamics, resulting in notable gains in motion fidelity, temporal stability, and rendering quality for online reconstruction.

Extensive experiments demonstrate that MoRGS delivers state-of-the-art reconstruction quality on dynamic scene datasets~\cite{n3dv,streamrf} and recovers true 3D dynamics across views and time through explicit per-Gaussian motion reasoning. To summarize, our main contributions are:

\begin{itemize}
\item 
We propose MoRGS, an efficient online framework for explicit per-Gaussian motion reasoning that leverages sparse motion cues to align Gaussian updates with true 3D scene dynamics.

\item 
We introduce flow-guided motion supervision that combines a per-Gaussian motion offset field with a per-Gaussian motion confidence, regularizing motion under sparse, view-limited supervision and prioritizing truly dynamic Gaussians to improve temporal consistency.

\item 
We demonstrate state-of-the-art rendering quality and motion fidelity across diverse dynamic scene benchmarks while maintaining streamable performance.

\end{itemize}
\section{Related Work}
\label{sec::related_work}

\subsection{Offline Dynamic Scene Reconstruction}

Building on advances in novel view synthesis (NVS), early progress on dynamic scene reconstruction was driven by NeRF-based methods~\cite{nerf}, where motion is commonly modeled as canonical-space deformations~\cite{dnerf,nerfies}. To improve efficiency while retaining fidelity, subsequent works extend radiance fields to four dimensions with structured parameterizations such as explicit grids, plane factorizations, and tensor decompositions~\cite{tensorf,kplanes,hexplane,tensor4d}. These designs accelerate optimization and inference compared to vanilla NeRF, yet they still rely on sequence-level objectives and typically require lengthy training.

More recently, 3DGS~\cite{3dgs} has become a mainstream approach for NVS due to its real-time rendering and fast training. Extending 3DGS to dynamic scenes, subsequent works~\cite{4dgs,4d-gs,rotor4dgs,fedgs,dn4dgs,spacetimegaussian} typically pair a canonical Gaussian field with spatiotemporal deformation fields to explain motion. Although these offline 4D Gaussian approaches achieve high quality by optimizing over long sequences with rich temporal constraints, this reliance makes them ill-suited for streaming. They assume full-sequence, non-causal training, and substantial computation, limiting their applicability when frames arrive sequentially and future observations are unavailable.

\subsection{Online Dynamic Scene Reconstruction}
Online approaches reconstruct scenes on-the-fly using only local temporal context, and they operate under strict constraints on training and rendering latency, bandwidth, and visual quality. Early NeRF-based attempts toward streamability optimized per-frame residuals or incrementally adapted the model~\cite{rerf,enerf,streamrf,inv}, but they suffered from slow convergence, high memory usage, and long training times.

Recent work predominantly adopts 3DGS for real-time rendering. A straightforward baseline, Dynamic3DGS~\cite{dynamic3dgs}, optimizes inter-frame residuals to track changes between consecutive frames~\cite{dynamic3dgs}. Building on this, several methods~\cite{3dgstream,queen,igs,hicom,4dgc,recon} targeting online scalability pursue low-latency encoding (training) and decoding (rendering) as well as low bandwidth usage and memory footprint. 3DGStream~\cite{3dgstream} accelerates training by modeling per-frame geometric transforms with a neural transformation cache. HiCoM~\cite{hicom} introduces hierarchical coherent motion modeling to reduce capacity and speed up optimization. QUEEN~\cite{queen} applies a quantization–sparsity scheme to cut per-frame storage and training cost. 4DGC~\cite{4dgc} designs a compression-friendly 4D Gaussian representation to improve streamability under tight storage. Instant Gaussian Stream~\cite{igs} proposes a generalizable anchor-based motion model that predicts motion in a single forward pass, thereby reducing per-frame optimization.

While these online methods achieve fast training and rendering with a low per-frame memory footprint, they primarily optimize for efficiency and rendering quality, leaving per-Gaussian motion under-constrained. In contrast, we explicitly reason about per-Gaussian motion by leveraging sparse motion cues, enabling streamable reconstruction with high motion fidelity and strong temporal consistency.

\begin{figure*}[h!]
\vspace{-25pt}
\begin{center}
    \includegraphics[width=0.96\textwidth]{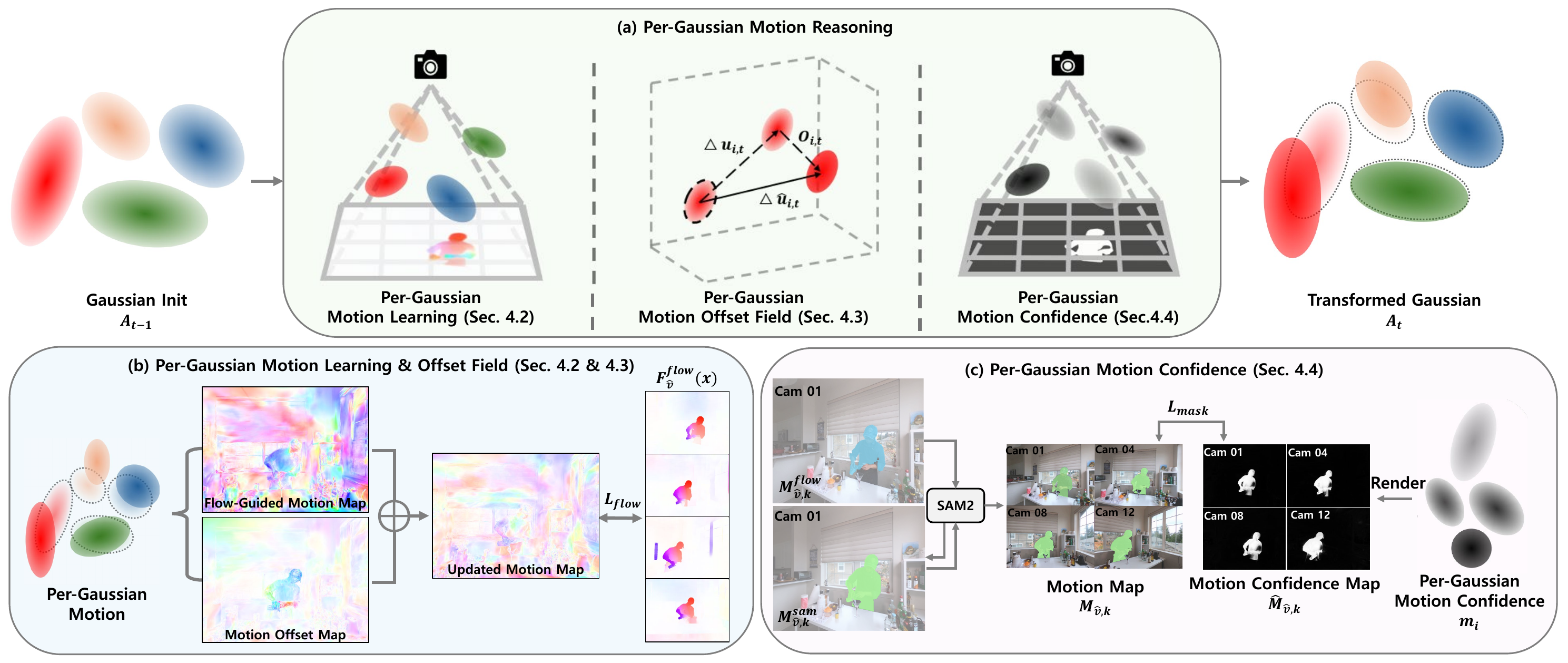}
    \vspace{-5pt}
    \captionof{figure}{
    \textbf{Illustration of the MoRGS framework.}
    (a) We incrementally update Gaussian attributes at each time step while jointly modeling per-Gaussian motion between frames.
    (b) Per-Gaussian motion is guided by sparse motion cues and refined by a per-Gaussian motion offset field to compensate for discrepancies in the sparse motion cues.
    (c) To identify dynamic Gaussians, we obtain motion masks by thresholding the motion cues and then apply a segmentation model for view consistency. The per-Gaussian motion confidence is learned from these masks to suppress redundant background motion, improve temporal consistency, and concentrate learning on large motions.
    }

\label{fig:framework}
\end{center}
\vspace{-20pt}
\end{figure*}
\section{Preliminary}
\label{sec:preliminary}

\subsection{3D Gaussian Splatting}

3DGS~\cite{3dgs} represents a 3D scene as a set of anisotropic Gaussians with attributes $\mathcal{A}=\{ \mu, S, R, o, c\}$, where each Gaussian is parameterized by geometry parameters $(\mu, S, R)$ for position, scale, and rotation, and appearance parameters $(o, c)$ for opacity and color. The color of a rendered pixel at image-plane location $x$ is obtained by point-based $\alpha$-blending of depth-sorted Gaussians as:
\begin{equation}
  C({x}) = \sum_{i=1}^{N}  T_i \, c_i \, \alpha_i , \qquad T_i = \prod_{j=1}^{i-1}(1-\alpha_j),
  \label{eq:alpha_blend}
\end{equation}
where $T_i$ and $\alpha_i$ denote the accumulated transmittance up to the $(i\!-\!1)^{th}$ Gaussian along the ray and the rendering weight induced by the opacity and projected Gaussian covariance.

Training proceeds end-to-end using differentiable rasterization and a reconstruction loss that combines $\mathcal{L}_1$ and D-SSIM losses as:
\begin{equation}
  \mathcal{L}_{\text{recon}} = (1 - \lambda)\mathcal{L}_1 + \lambda \mathcal{L}_{\text{D-SSIM}},
  \label{eq:training_loss}
\end{equation}
where $\lambda$ balances structural and photometric fidelity. This objective jointly optimizes all Gaussian attributes $\mathcal{A}$ to recover high-fidelity geometry and appearance from multi-view images.

\subsection{Online Gaussian Attribute Modeling}

For dynamic scene reconstruction, online frameworks extend 3DGS by modeling inter-frame differences through attribute residuals. At time step $t$, the Gaussian attributes $\mathcal{A}_t$ are obtained by adding learnable residuals $\mathcal{R}_t$ to the previously optimized attributes $\mathcal{A}_{t-1}$:
\begin{equation}
    \mathcal{A}_t = \mathcal{A}_{t-1} + \mathcal{R}_t,
    \label{eq:residual}
\end{equation}
where $\mathcal{R}_t$ captures per-Gaussian temporal variations in geometry and appearance. This causal recurrence supports sequential, adaptive updates as new frames arrive, enabling real-time training and rendering without access to the full sequence.

\section{Method}
\subsection{Overview}

Given a multi-view video $\{I_{v,t}\}$ with view index $v \in \mathcal{V}$ and time step $t \in [0, +\infty)$,  our goal is to estimate per-Gaussian motion that faithfully follows scene dynamics in an online setting. First, we compute motion cues from a sparse set of views and use them to guide per-Gaussian motion (Sec.~\ref{sec:method:flowsupervision}). Second, to account for errors arising from sparse, view-limited supervision, we introduce a per-Gaussian motion offset field that corrects flow-induced misalignments and enforces 3D geometric consistency (Sec.~\ref{sec:method:motionoffset}). Third, we learn a per-Gaussian motion confidence that weights residual updates of Gaussian attributes, emphasizing Gaussians with reliable motion evidence while suppressing updates for nearly static ones (Sec.~\ref{sec:method:motionconfidence}). An overview of the framework is shown in Fig.~\ref{fig:framework}.

\subsection{Per-Gaussian Motion Learning}
\label{sec:method:flowsupervision}
Optimizing inter-frame transformations using only photometric loss induces geometry-inconsistent per-Gaussian motion, causing Gaussians to chase appearance changes rather than true scene motion. To guide motion toward the scene’s actual dynamics, we align each Gaussian’s projected 3D displacement with pixel-level motion cues.

\noindent\textbf{Efficient 2D Motion Cues.}
Optical flow is a widely used representation of pixel-level motion, and we estimate it using a pretrained network $F^{\text{flow}}$, which serves as a low-cost 2D motion prior providing directional supervision for per-Gaussian motion. However, computing flow for all views is impractical for online training. We therefore compute flow between consecutive frames on a subset of key views $\hat{v}$ as:
\begin{equation}
 F_{\hat{v}}^{\text{flow}}(x)=F^{\text{flow}}(I_{\hat{v},t}, I_{\hat{v},t-1}).
\label{eq:flow_loss}
\end{equation}

\noindent\textbf{Pixel-Gaussian Motion Correspondence.}
To leverage the image-space motion cues $F_{\hat{v}}^{\text{flow}}({x})$, we project each Gaussian’s 3D displacement and render a per-view Gaussian motion map with the differentiable Gaussian renderer. For each Gaussian, the 3D motion is defined as the difference between its updated position in the current frame ${\mu}_{i,t}$ and its optimized position in the previous frame ${\mu}_{i,t-1}$ as:
\begin{equation}
\Delta {\mu}_{i,t} =  ({\mu}_{i,t} - {\mu}_{i,t-1}).
\label{eq:gaussian motion}
\end{equation}

We project this 3D displacement to the image plane of view $\hat{v}$ using a first-order linearization $\pi_{\hat{v}}$ of the camera projection, yielding a 2D Gaussian motion vector. This projected motion is then rendered as a per-pixel Gaussian motion map through normalized $\alpha$-blending:
\begin{equation}
F_{\hat{v}}^G({x}) = \sum_{i=1}^{N} w_i({x}) \, \pi_{\hat{v}}(\Delta {\mu}_{i,t}), \label{eq:gaussian_flow_render} 
\end{equation}
where $w_i({x}) = \frac{T_i({x}) \, \alpha_i}{\sum_i T_i({x}) \, \alpha_i} $ denotes transmittance-weighted opacities normalized along the ray, ensuring volumetric consistency of the rendered Gaussian motion.

We align the rendered Gaussian motion map $F_{\hat{v}}^G(x)$ with the optical flow $F_{\hat{v}}^{flow}({x})$ from key views and train using an endpoint-error loss as:
\begin{equation}
\mathcal{L}_{\text{flow}} =
\sum_{\hat{v}} \| F_{\hat{v}}^{\text{flow}}({x}) - F_{\hat{v}}^G({x}) \|_2.
\label{eq:flow_loss}
\end{equation}
This flow-guided motion learning constrains per-Gaussian displacements to follow the observed 2D motion while remaining fully differentiable within the Gaussian rendering framework, thereby reducing appearance-driven motion updates.

\subsection{Per-Gaussian Motion Offset Field}
\label{sec:method:motionoffset}
Flow-guided supervision provides reliable directional cues for Gaussian motion, especially around large displacements on key views. To further promote a coherent 3D motion field across views and time, we complement it with a learnable per-Gaussian motion offset field  $\{O_{i,t}\}_{i=1}^N$ with $O_{i,t}\in\mathbb{R}^{3}$, which acts as a spatial refinement term during optimization. As a learnable offset attached to each Gaussian and continuously updated from all of its observations, it aggregates multi-view evidence and mitigates the view-dependence of sparse motion cues. Unlike the rendered Gaussian motion map, which blends contributions from multiple Gaussians along each ray and can blur attribution, the offset directly updates individual Gaussians, correcting inconsistent or noisy flow signals and stabilizing motion learning.

Recall that the per-Gaussian 3D motion between frames is defined as $\Delta \mu_{i,t}$ in Eq.~\eqref{eq:gaussian motion}. 
Under the flow supervision in Sec.~\ref{sec:method:flowsupervision}, this term becomes a flow-guided base estimate of per-Gaussian motion. 
To refine this base motion under 3D constraints, we separately model a learnable per-Gaussian offset and rewrite the Gaussian motion as:
\begin{equation}
\Delta \hat{\mu}_{i,t} = \underbrace{\Delta {\mu}_{i,t}}_{\text{flow-guided}} + \underbrace{O_{i,t}}_{\text{learnable offset}}\!\!\!\!\!.
\label{eq:gaussian motion2}
\end{equation}
By separately modeling the per-Gaussian motion offset field, the flow-guided base motion $\Delta\mu_{i,t}$ encodes the displacement suggested by image-space motion cues on sparse key views, whereas the offset $O_{i,t}$ is optimized from all views that observe the Gaussian and thus aggregates multi-view gradients to refine this displacement. For cues that are consistent across views and time, the offset remains small and the base motion dominates. In contrast, if the cues disagree with the underlying 3D geometry, the offset compensates for the discrepancy so that the effective displacement follows the scene’s true motion rather than overfitting to sparse flow signals. We regularize the offset field to keep corrections small and prevent them from overpowering the flow-guided base motion with $\mathcal{L}_{\text{off}}=\bigl\| O_{i,t}\bigl\|_1$.

\begin{figure}[t] 
  \centering
  \includegraphics[width=\columnwidth]{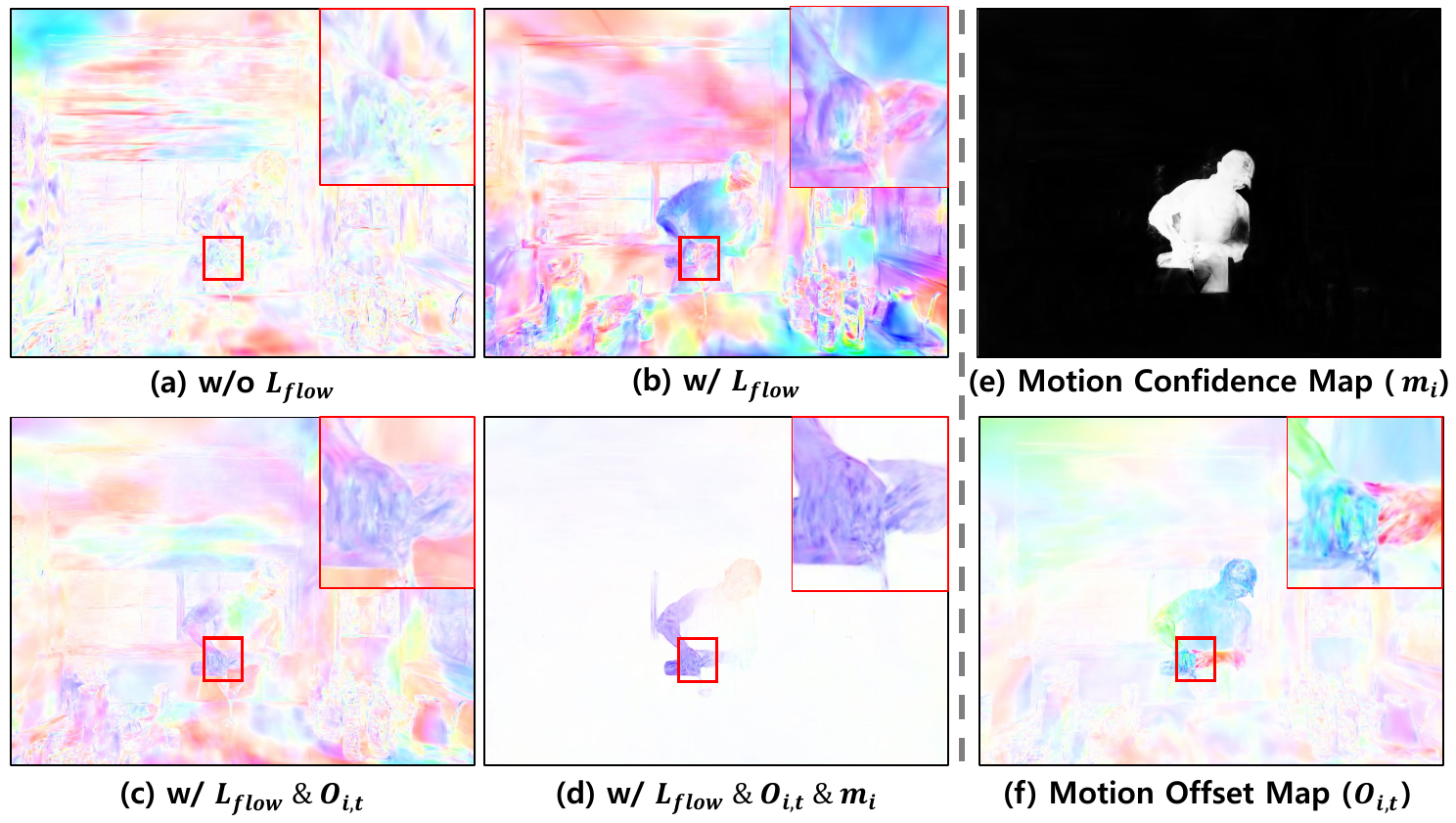} 
\caption{\textbf{Per-Gaussian Motion Visualization.}
We visualize per-Gaussian motion under (a) no flow guidance,
(b) sparse flow guidance only,
(c) sparse flow guidance with motion offset,
and (d) sparse flow guidance with both motion offset and motion confidence,
while (e) and (f) show the learned motion confidence map and motion offset map, respectively.
}
  \label{fig:map}
\vspace{-10pt}
\end{figure}

In Fig.~\ref{fig:map} (a)–(c), (f), we show per-Gaussian motion under different supervision schemes. Without flow guidance (a), per-Gaussian motions drift largely independently of the true dynamic of the scene. Adding sparse flow guidance (b) restores directional consistency but still propagates spurious motion to incorrect Gaussians. Incorporating the motion offset (c) suppresses these artifacts and yields smoother, geometry-consistent motion. The motion-offset map (f) shows that $O_{i,t}$ concentrates where flow supervision misguides Gaussians, effectively absorbing spurious displacements.

\begin{table*}[!t]
\footnotesize
\centering
\setlength{\tabcolsep}{2.5pt}
\begin{tabular}{l|l|ccccccc}
\toprule
Category & Method & PSNR (dB)$\uparrow$ & SSIM$\uparrow$ & LPIPS$\downarrow$ & Storage (MB)$\downarrow$ & Training (sec)$\downarrow$ & Render (FPS)$\uparrow$\\
      \midrule
\multirow{4}{*}{Offline}
& Real-Time 4DGS~\cite{4d-gs}     & 32.01 &  \cellbest{0.972} &  0.055  &  3.30 & 108 & \cellsecond{114} \\
& 4DGS~\cite{4d-gs}     & 31.02 &  \cellsecond{0.940} &  0.150  &  \cellbest{0.30} & \cellsecond{10} & 30 \\
& STG~\cite{spacetimegaussian} & \cellsecond{32.05} & \cellbest{0.972} & \cellsecond{0.044} & 0.67 & 120 & 110 \\
& Swift4D~\cite{swift4d}       &  \cellbest{32.23} & \cellbest{0.972} & \cellbest{0.043} &  \cellsecond{0.40} &  \cellbest{5.0} & \cellbest{125} \\

\midrule
\multirow{7}{*}{Online}
% & Dynamic 3DGS~\cite{dynamic3dgs} & 30.67 & 0.931 & - & 9.2 & - & 460 \\ %4DGC
& StreamRF~\cite{streamrf}     & 30.68 & - & - & 31.5 & 15 & 8.3 \\ %QUEEN
%& IGS$^\ast$~\cite{igs}                & 34.15 & - & - & 2.5 & 2.1 & 100 \\ %QUEEN
& 3DGStream~\cite{3dgstream} & 31.67 & - &  - & 7.80 & 13 & \cellsecond{215} \\ %QUEEN
& HiCoM~\cite{hicom}           & 31.17 &  - &  - & 0.90 &  11 &  \cellbest{260} \\ %HICOM PAPER
& QUEEN-l~\cite{queen}         & 32.19 & \cellsecond{0.946} & 0.136 & \cellsecond{0.75} & \cellbest{2.9} &  186 \\ %QUEEN
& 4DGC~\cite{4dgc}             & 31.58 & 0.943 & -  & \cellbest{0.50} & 50 & 168 \\ %4DGC
\cmidrule(lr){2-8} 
& MoRGS-s (ours)            & \cellsecond{32.42} & \cellbest{0.950} & \cellsecond{0.119} & 3.79 & \cellsecond{3.4} & \cellsecond{215} \\
& MoRGS-l (ours)           & \cellbest{32.53} & \cellbest{0.950} &  \cellbest{0.118}  & 3.80 & 4.0 & 200 \\

      \bottomrule
    \end{tabular}
\caption{
\textbf{Quantitative comparison} on the N3DV dataset. For each scene, all metrics are averaged over 300 frames. Storage and training time both include the initial frame size and time.  Red and orange highlight the best and second-best results in each category, respectively.}
\label{tab:benchmark1}
\vspace{-15pt}
\end{table*}

\subsection{Per-Gaussian Motion Confidence}
\label{sec:method:motionconfidence}
Effective online reconstruction hinges on prioritizing updates to truly dynamic Gaussians while leaving static ones largely unchanged to preserve temporal consistency. We therefore introduce a per-Gaussian motion confidence $m_i \in [0,1]$ that acts as a motion likelihood and weights Gaussian attribute residuals, suppressing updates for nearly static Gaussians while emphasizing dynamic ones during training. Together with flow-guided motion learning and the motion offset field, this confidence yields motion updates that remain temporally consistent in static regions while concentrating on Gaussians with reliable motion evidence.

\noindent\textbf{2D Motion Segmentation Mask.}
Starting from the initial frame, we compute 2D binary motion masks on periodically sampled keyframes $K$. These masks supervise the per-Gaussian motion confidence $m_i$ to distinguish dynamic from static Gaussians. For each $k \in K$, we obtain a per-pixel motion mask from optical flow estimated on a subset of key views (Sec.~\ref{sec:method:flowsupervision}). Under the static-camera assumption in multi-view capture, pixels with negligible flow magnitude are treated as static, whereas those whose magnitude exceeds a threshold $\lambda^{\text{flow}}$ are labeled as dynamic as:
\begin{equation}
M_{\hat{v},k}^{\text{flow}} = \| F^{\text{flow}}(I_{\hat{v},k}, I_{\hat{v},k-1}) \| > \lambda^{\text{flow}}.
\label{eq:raft_mask}
\end{equation}

Although flow-based masks $M_{\hat{v},k}^{\text{flow}}$ capture large motions, they are strongly view-dependent and can label the same 3D region as dynamic in some views and static in others, which is problematic when supervising a single scalar confidence $m_i$ shared by each Gaussian across all views. We therefore refine these masks into view-consistent, object-level regions using SAM2~\cite{ravi2024sam}. For each key view $\hat{v}$ at keyframe $k$, we feed the RGB frame $I_{\hat{v},k}$ and the flow mask $M_{\hat{v},k}^{\text{flow}}$ into the segmentation model $F^{\text{sam}}$ to obtain an object-level mask $M_{\hat{v},k}^{\text{sam}}$. We then propagate the key view mask to the remaining views. The final key view motion masks are obtained by fusing the two masks as:

\begin{equation}
M_{\hat{v},k}^{\text{sam}} = F^{\text{sam}}(I_{\hat{v},k},M_{\hat{v},k}^{\text{flow}}), \ \ 
M_{\hat{v},k} = M_{\hat{v},k}^{\text{flow}} \cup M_{\hat{v},k}^{\text{sam}}.
\label{eq:mask_propagation}
\end{equation}
This recovers missed motion regions and maintains object-consistent boundaries, while avoiding costly per-frame computation by restricting SAM2 inference to key views and keyframes only.

\noindent\textbf{Mask-Guided Confidence Learning.}
The aggregated motion masks $M_{\hat{v},k}$ are then lifted into the Gaussian domain by learning a single-channel motion confidence $m_i$ for each Gaussian. We obtain $m_i$ through sigmoid activation, so it represents the likelihood that the $i^{th}$ Gaussian is dynamic. Consistent with the $\alpha$-blending rule in Eq.~\eqref{eq:alpha_blend}, the rendered motion confidence map is computed by blending per-Gaussian motion confidence along each ray as $\tilde{M}_{\hat{v},k}(\mathbf{x}) = \sum_{i=1}^{N}  T_i \, m_i \, \alpha_i $ and optimized by motion mask $M_{\hat{v},k}$ as:
\begin{equation}
\mathcal{L}_{\text{mask}} = \sum_{\hat{v}} \bigl\| \tilde{M}_{\hat{v},k} - M_{\hat{v},k} \bigr\|_1.
\label{eq:motion_loss}
\end{equation}

The learned per-Gaussian motion confidence $m_i$ is integrated into the residual update process of Gaussian attributes (Eq.~\eqref{eq:residual}):
\begin{equation}
\mathcal{A}_{i,t} = \mathcal{A}_{i,t-1} + m_i \odot \mathcal{R}_{i,t},
\label{eq:motion_gated_residual}
\end{equation}
where $\odot$ denotes element-wise weighting across attribute dimensions. The learnable motion confidence $m_i$ down-weights updates for nearly static components and prioritizes dynamic ones, which suppresses redundant background motion and leads to more temporally consistent reconstructions, while also accelerating the modeling of large motions, as shown in Fig~\ref{fig:map} (d) and (e).

\subsection{Online End-to-End Training}
\label{sec:method:training}

\noindent\textbf{Initial Frame Reconstruction.}
We initialize Gaussians from SfM points~\cite{sfm} and set the motion confidence $m_i$ to zero. Following 3DGS~\cite{3dgs}, we first optimize static attributes while freezing motion confidence until densification, preventing motion-gradient interference and preserving early quality. After densification, we enable motion confidence and apply $\mathcal{L}_{\text{mask}}$ on key views to guide subsequent motion updates. For the first frame only, the motion mask is computed from the next frame due to the lack of a prior frame.

\noindent\textbf{Sequential Frame Reconstruction.} 
After reconstructing the initial frame, we process frames sequentially by updating Gaussian attribute residuals from the previous frame. We jointly optimize the per-Gaussian motion offset and confidence as:
\begin{equation}
\mathcal{L}_{\text{total}} =
\mathcal{L}_{\text{recon}} 
+ \lambda_{\text{mask}} \mathcal{L}_{\text{mask}} 
+ \lambda_{\text{flow}} \mathcal{L}_{\text{flow}} 
+ \lambda_{\text{off}} \mathcal{L}_{\text{off}},
\label{eq:final_motion_loss}
\end{equation}
where $\lambda_{\text{mask}}$, $\lambda_{\text{flow}}$, and $\lambda_{\text{off}}$ balance motion reasoning and reconstruction quality. Note that we compute $\mathcal{L}_{\text{mask}}$ and $\mathcal{L}_{\text{flow}}$ only on key views, and $\mathcal{L}_{\text{mask}}$ is further restricted to keyframes. We also perform the 3DGS densification stage at each time step to capture newly observed or finer scene content.

\section{Experiments}
%\vspace{-5pt}

\begin{table}[t]
\footnotesize
\centering
\setlength{\tabcolsep}{3.5pt}
\begin{tabular}{l|cccccc}
\toprule
\cmidrule(lr){2-7} 
& PSNR  & SSIM & LPIPS & Storage & Training & Render \\
& (dB)$\uparrow$ & $\uparrow$ & $\downarrow$ & (MB)$\downarrow$  & (sec)$\downarrow$  & (FPS)$\uparrow$ \\
\midrule
3DGStream~\cite{3dgstream}    & 30.79 &  - &  - & 4.10 & 7.2 & 288   \\                 
QUEEN-l$^\dag$~\cite{queen}      & 29.47 & 0.946 & 0.185 & \textbf{0.38} & \textbf{1.5} & \textbf{317}    \\                               
HiCoM~\cite{hicom}        & 26.73 &  - &  - & 0.60 & 9.8 & 258   \\                                
4DGC~\cite{4dgc}         & 28.08 &  0.922 &  - & 0.42 & 33.8 & 213 \\                             
\cmidrule(lr){0-6} 
MoRGS (ours)                & \textbf{31.79} &  \textbf{0.957} &  \textbf{0.152} & 2.01 & 2.3 & 308  \\                 

\bottomrule
\end{tabular}
\caption{
\textbf{Quantitative comparison} on Meet Room dataset. QUEEN-l$^\dag$ refers to our re-implementation result through official code in the same experimental environment.}
\label{tab:benchmark2}
\vspace{-15pt}
\end{table}

\begin{figure*}[h!]
\maketitle
\begin{center}
    \includegraphics[width=0.96\textwidth]{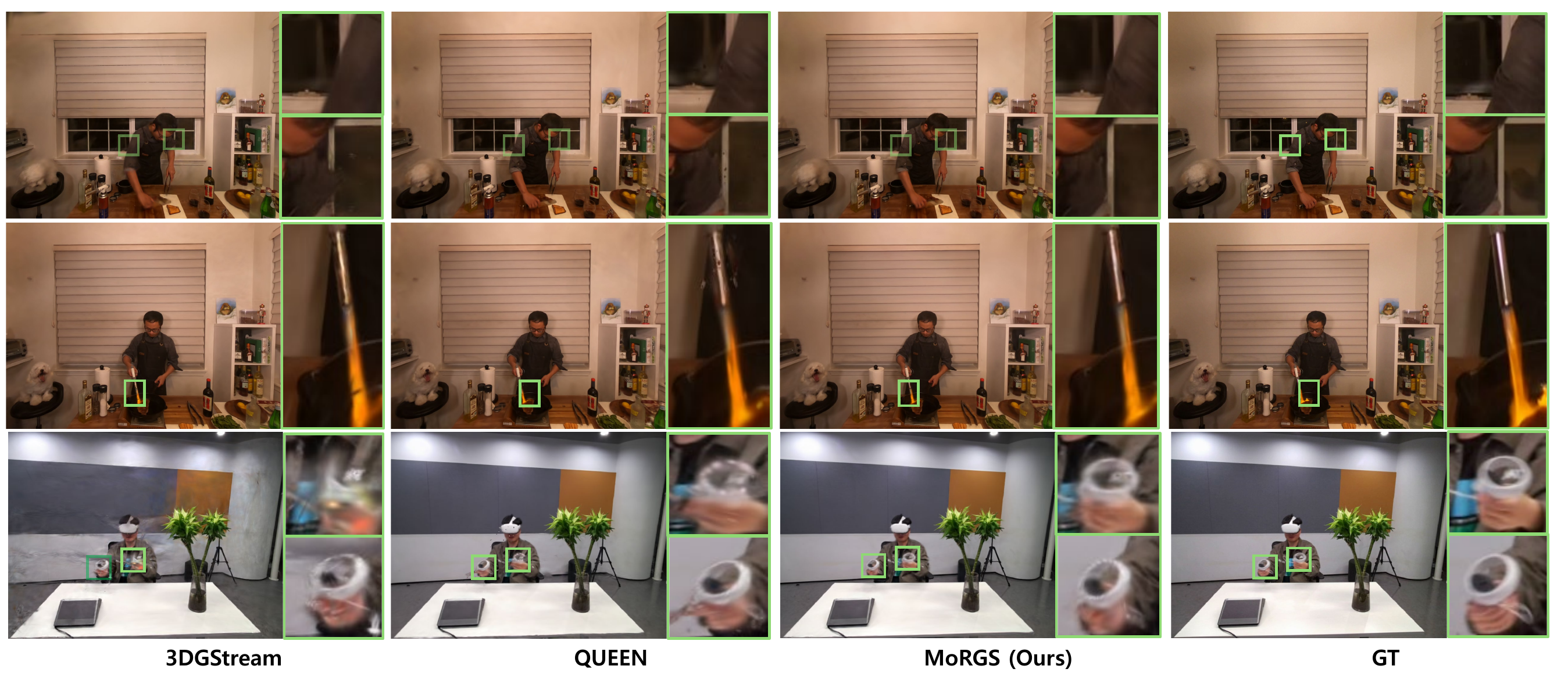}
    \vspace{-5pt}
    \captionof{figure}{
        \textbf{Qualitative Results.} A visualization of various scenes in N3DV and Meet Room dataset. We include additional video results in the supplementary material. 
    }
\label{fig:quality}
\end{center}
\vspace{-10pt}
\end{figure*}

\begin{figure*}[h!]
\maketitle
\begin{center}
    \includegraphics[width=0.96\textwidth]{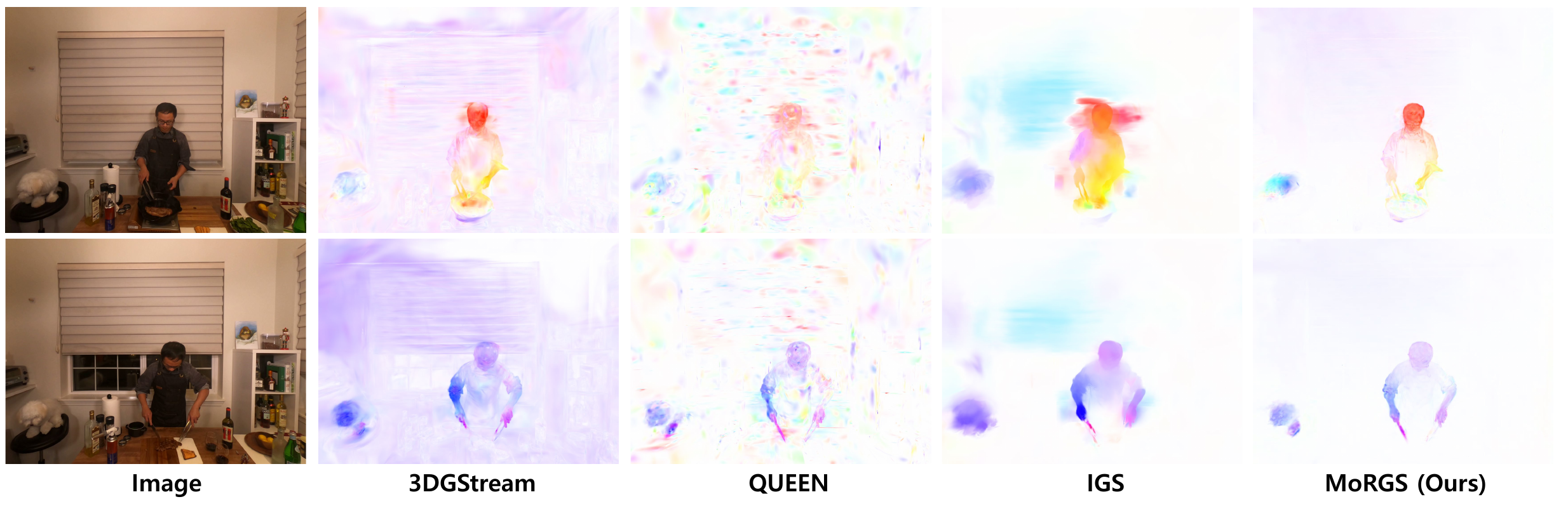}
    \vspace{-5pt}
    \captionof{figure}{
        \textbf{Gaussian Motion Visualization.} We observe that motion updates are confined to genuinely dynamic regions, and the recovered per-Gaussian motion closely follows the true scene dynamics.}

\label{fig:motion_quality}
\end{center}
\vspace{-10pt}
\end{figure*}

\subsection{Datasets and Implementation Details}
\label{sec:experiment:setup}
We evaluate our method on two dynamic scene datasets. (1) Neural 3D Videos (N3DV)~\cite{n3dv} consists of six indoor scenes with forward-facing 20-view videos. (2) Meet Room~\cite{streamrf} includes three indoor scenes with 13-view videos. All sequences contain 300 frames at 30 FPS, and the central view is held out for testing.

Our MoRGS reconstructs the initial frame for 10k and 15k iterations on N3DV and Meet Room, respectively, with the per-Gaussian motion confidence enabled only during the final 2k iterations. Subsequent frames are optimized online for 8 epochs per view. Optical flow is computed using SEA-RAFT~\cite{searaft} on 4 designated views per scene, and keyframes are sampled every 5 frames. All experiments are conducted on a single NVIDIA RTX A5000 GPU. Further implementation details are provided in the supplementary material.

\subsection{Experiment Results}
\label{sec:experiment:results}
\textbf{Quantitative comparisons.}
We compare our method with state-of-the-art online approaches on two datasets in Tab.~\ref{tab:benchmark1} and Tab.~\ref{tab:benchmark2}, using PSNR, SSIM, LPIPS, storage size, training time, and render speed as metrics. For completeness, Tab.~\ref{tab:benchmark1} also reports offline methods. For statistical reliability and fairness, training time and rendering speed are re-evaluated on the same GPU as ours. In addition, we report a lightweight variant, MoRGS-s, where the per-frame residuals are trained for only 5 epochs per view.

As shown in Tab.~\ref{tab:benchmark1}, our method achieves state-of-the-art rendering quality while maintaining comparable training latency to online baselines. While QUEEN trains slightly faster ($-1.1$~s), it yields lower quality ($-0.34$~dB). 4DGC attains the smallest storage but requires much longer training ($+46$~s). The small variant MoRGS-s reduces per-frame training time by 0.6~s over our full model while retaining the best PSNR among online methods. We also evaluate on Meet Room in Tab.~\ref{tab:benchmark2}. Our method consistently achieves the best quality across diverse motion patterns while maintaining similar latency. These improvements underscore the effectiveness of per-Gaussian motion modeling, which aligns each Gaussian's motion update with the underlying scene motion.

To substantiate that Gaussian updates in our method are confined to genuinely dynamic regions, we also report masked Total Variation (mTV) measured within predefined static-region masks (Tab.~\ref{tab:mTV}). Our method achieves the lowest mTV among 3DGStream~\cite{3dgstream} and QUEEN~\cite{queen}, indicating stronger temporal stability in static areas.

\begin{figure}[t] 
  \centering
  \includegraphics[width=\columnwidth]{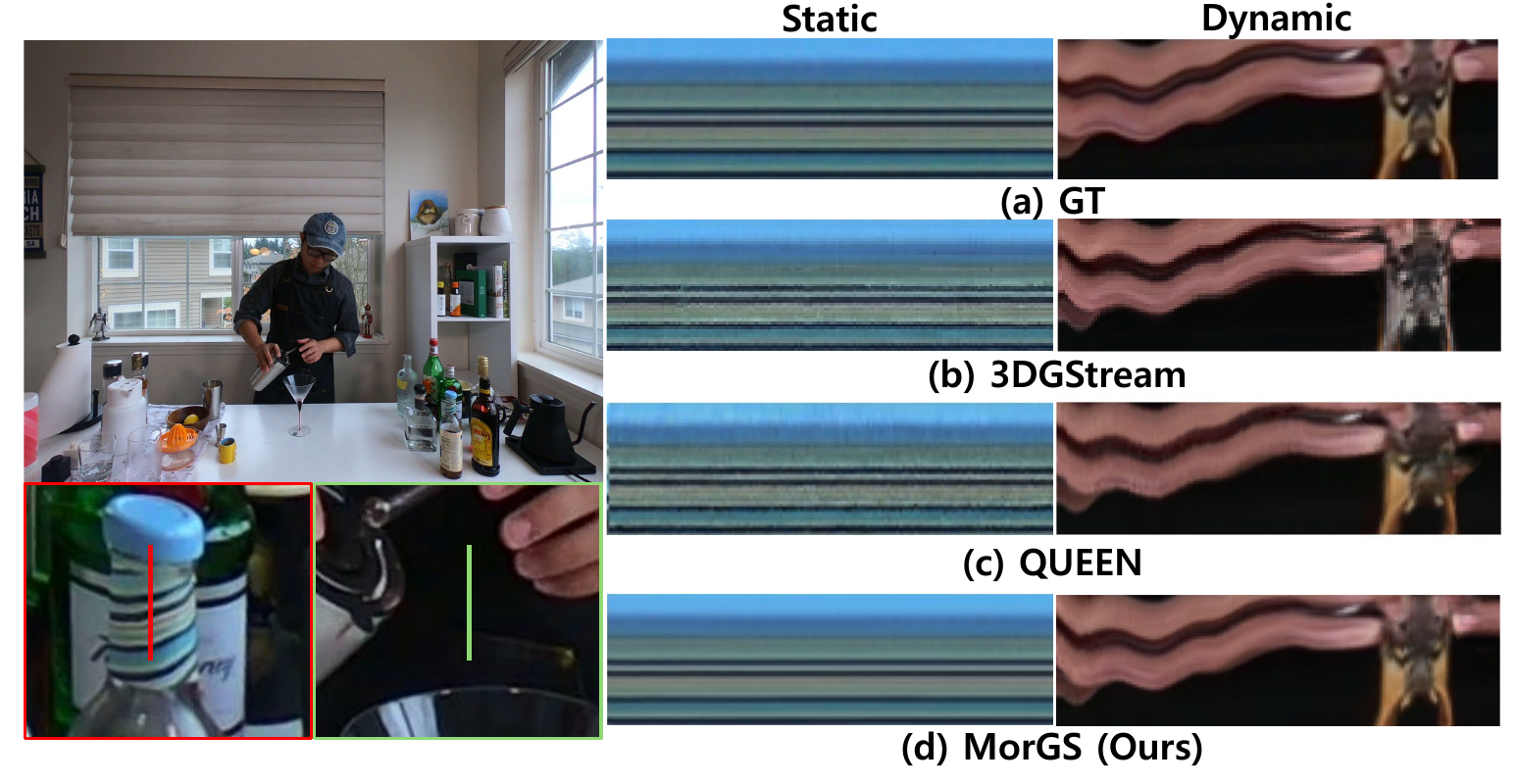} % or \linewidth
  \vspace{-20pt}
  \caption{\textbf{Temporal Consistency Comparison.} A visualization of spatiotemporal images from a fixed vertical scanline over time. }
  \label{fig:spatiotemporal}
%\vspace{-5pt}
\end{figure}

% A visualization of spatiotemporal images built from a fixed vertical scanline over time.
\begin{table}[t]
    \centering
    \footnotesize
    \setlength{\tabcolsep}{4pt}
    \begin{tabular}{l l ccc}
    \toprule
    Scene & Method  & PSNR (dB)$\uparrow$ & SSIM $\uparrow$ & $\mathrm{mTV}_{\!\times\!100} \downarrow$ \\
    \midrule
    \multirow{3}{*}{\makecell[c]{COFFEE\\MARTINI}}
        & 3DGStream               & 27.75 & -  & 0.892 \\
        & QUEEN-l                   & 28.38 & 0.915  & 1.51 \\
        & Ours                    & \textbf{29.01} & \textbf{0.923}  & \textbf{0.671} \\
    \cmidrule(lr){1-5}
    \multirow{3}{*}{\makecell[c]{FLAME\\STEAK}}
        & 3DGStream               & 28.39 & -  & 0.668 \\
        & QUEEN-l                   & 29.25 & 0.923  & 1.30 \\
        & Ours                    & \textbf{30.11} & \textbf{0.931}  & \textbf{0.587} \\

    \bottomrule
    \end{tabular}
    \caption{Analysis of rendering quality and temporal consistency on two N3DV scenes. mTV is computed only in static regions defined by predefined masks.}

    \label{tab:mTV}
\vspace{-15pt}
\end{table}

\textbf{Qualitative comparisons.}
As illustrated in Fig.~\ref{fig:quality}, our approach reconstructs dynamic scene details more completely by explicitly modeling per-Gaussian motion, surpassing 3DGStream~\cite{3dgstream} and QUEEN~\cite{queen} across datasets. By learning true scene dynamics rather than relying on image-space proxies, it robustly captures large motions (e.g., torch, hands) while reducing flicker and other artifacts in static regions.

In Fig.~\ref{fig:motion_quality}, we visualize Gaussian motion updates. Whereas 3DGStream~\cite{3dgstream}, QUEEN~\cite{queen}, and IGS~\cite{igs} produce redundant updates in static areas or respond primarily to photometric changes, our method updates only Gaussians with high motion confidence and aligns their displacements with the underlying 3D motion, yielding consistent updates across time and views.

Furthermore, Fig.~\ref{fig:spatiotemporal} presents spatiotemporal images obtained by horizontally concatenating a fixed vertical line over time. 3DGStream~\cite{3dgstream} and QUEEN~\cite{queen} produce noisy spatiotemporal patterns with artifacts in both static and dynamic regions, indicating temporal inconsistency. In contrast, our method improves the temporal consistency in online reconstruction, producing sharper and more stable spatiotemporal images. Additional qualitative results are provided in the supplementary material.

\subsection{Ablation Study}
\label{sec:experiment:ablation}

As shown in Tab.~\ref{tab:ablation}, we ablate three components on the N3DV and Meet Room dataset, including per-Gaussian motion learning (ML), motion offset (MO), and motion confidence (MC).

\noindent\textbf{Effect of Per-Gaussian Motion Learning}
Starting from the baseline, guiding per-Gaussian motion with sparse motion cues improves PSNR by $+0.52$~dB on N3DV and $+1.15$~dB on Meet Room, while incurring only a $0.4$~s increase in per-frame training time for optical flow, since computation is restricted to a subset of views. As shown in Tab.~\ref{tab:motion_offset_views}, increasing the number of supervised views strengthens motion supervision and further improves accuracy, but also raises the per-frame training time, revealing a clear accuracy–efficiency trade-off.

\noindent\textbf{Effect of Per-Gaussian Motion Offset}
The per-Gaussian motion offset further improves PSNR by $+0.36$~dB and $+0.66$~dB on N3DV and Meet Room, respectively, on top of motion learning. By refining motion directly in 3D space, it compensates for errors from sparse flow-guided supervision and enforces consistency across views and time. In Tab.~\ref{tab:motion_offset_views}, we observe that the benefit of the motion offset becomes more pronounced as motion supervision becomes sparser. Notably, using the offset with only four supervised views outperforms using eight supervised views without it, indicating that the motion offset enables more effective use of sparse motion cues.

\noindent\textbf{Effect of Per-Gaussian Motion Confidence}
Learning a per-Gaussian motion confidence further improves PSNR by $+0.32$~dB and $+0.58$~dB on N3DV and Meet Room, respectively. By prioritizing updates for Gaussians with high motion confidence, it suppresses redundant updates in static regions, leading to more temporally consistent reconstructions, and accelerates motion learning for dynamic Gaussians under large inter-frame motion.

\newcommand{\cmark}{\ding{51}} % ✓
\newcommand{\xmark}{\ding{55}} % ✗

\begin{table}[t]
    \centering
    \resizebox{\linewidth}{!}{%
    \setlength{\tabcolsep}{4pt}
    \begin{tabular}{ccc|ccc|ccc}
    \toprule
    \multicolumn{3}{c|}{Modules} & \multicolumn{3}{c|}{N3DV} & \multicolumn{3}{c}{Meet Room} \\
    ML & MO & MC & PSNR$\uparrow$ & SSIM$\uparrow$ & Train (s)$\downarrow$ & PSNR$\uparrow$ & SSIM$\uparrow$ & Train (s)$\downarrow$ \\
    \midrule
    \xmark & \xmark & \xmark & 31.33 & 0.946 & 3.3 & 29.40 & 0.952 & 1.7 \\
    \cmark & \xmark & \xmark & 31.85 & 0.948 & 3.7 & 30.55 & 0.954 & 2.1 \\
    \cmark & \cmark & \xmark & 32.21 & 0.949 & 3.8 & 31.21 & 0.955 & 2.2 \\
    \cmark & \cmark & \cmark & 32.53 & 0.950 & 4.0 & 31.79 & 0.957 & 2.3 \\

    \bottomrule
    \end{tabular}
    }
    \caption{Ablation on main components of our MoRGS framework.}
    \label{tab:ablation}

\end{table}

\begin{table}[t]
\centering
\scriptsize
\setlength{\tabcolsep}{4pt}
\begin{tabular}{c|cccc}
\toprule
 & \multicolumn{4}{c}{\# of Motion Supervision Views} \\
\cmidrule(lr){2-5}
Method & 2 & 4 & 8 & 12 \\
\midrule
w/o Motion Offset & 31.60     & 31.95   & 32.07   & 32.12   \\
w/ Motion Offset  & 31.82   & 32.21   & 32.22   & 32.22    \\
\midrule
Per-frame time (s) & 3.57 & 3.95 & 4.58 & 6.32 \\
\bottomrule
\end{tabular}
\caption{Ablation on number of motion supervision and offset in N3DV dataset.}
\label{tab:motion_offset_views}
\vspace{-15pt}
\end{table}

\section{Conclusion}
\label{sec::conclusion}
We presented MoRGS, an efficient per-Gaussian motion reasoning framework for online dynamic scene reconstruction. By leveraging sparse flow-based motion cues, a 3D motion offset field, and per-Gaussian motion confidence, our method explicitly models per-Gaussian motion and confines updates to truly dynamic Gaussians. Experiments on multiple dynamic scenes show that MoRGS achieves state-of-the-art rendering quality and motion fidelity among online methods while maintaining streamable performance.

\noindent\footnotesize\textbf{Acknowledgement.} This work was supported by the National Research Foundation of Korea (NRF) grant funded by the Korea government (MSIT)(No. RS-2024-00340745) and  Electronics and Telecommunications Research Institute (ETRI) grant funded by the Korean government [26ZC1100, Development of Spatial Media Technology and Interaction Technology for Convergence of the Real and Virtual World].

\clearpage

{
    \small
    \bibliographystyle{ieeenat_fullname}
    \bibliography{main}
}

\end{document}